# Efficient On-Chip Implementation of 4D Radar-Based 3D Object Detection on Hailo-8L

**Woong-Chan Byun**
M.S. student, CCS Graduate School of Mobility, KAIST
193 Munji-ro, Yuseong-gu, Daejeon 34051, Republic of Korea
+82-42-350-1285, woongchan.byun@kaist.ac.kr
**Dong-Hee Paek**
Ph. D. student, CCS Graduate School of Mobility, KAIST
193 Munji-ro, Yuseong-gu, Daejeon 34051, Republic of Korea
+82-42-350-1285, donghee.paek@kaist.ac.kr
**Seung-Hyun Song**
M.S. student, Graduate School of Advanced Security Science and Technology, KAIST
193 Munji-ro, Yuseong-gu, Daejeon 34051, Republic of Korea
+82-42-350-1285, shyun@kaist.ac.kr
**Seung-Hyun Kong***
Professor, CCS Graduate School of Mobility, KAIST
193 Munji-ro, Yuseong-gu, Daejeon 34051, Republic of Korea
+82-42-350-1285, skong@kaist.ac.kr

## ABSTRACT

4-Dimensional (4D) Radar have attracted attention in autonomous driving due to their ability to enable robust 3D object detection even under adverse weather conditions. To practically deploy such technologies, it is essential to achieve real-time processing within low-power embedded environments. Addressing this, we present the first on-chip implementation of a 4D Radar-based 3D object detection model on the Hailo-8L AI accelerator. Although conventional 3D convolutional neural network (CNN) architecture requires 5D inputs, the Hailo-8L only supports 4D tensors, posing a significant challenge. To overcome this limitation, we introduce a tensor transformation that reshapes 5D inputs into 4D formats during the compilation process, enabling direct deployment without altering the model structure. The proposed system achieves 46.47% $AP_{3D}$ and 52.75% $AP_{BEV}$, maintaining comparable accuracy to GPU-based models, while achieving an inference speed of 13.76 Hz. These results demonstrate the applicability of 4D Radar-based perception technologies to autonomous driving systems.

## INTRODUCTION

4D Radar have emerged as a key sensor in autonomous driving due to their robust object detection capabilities even under adverse weather conditions. (1) To practically deploy 4D Radar-based 3D object detection in autonomous systems, it is essential to ensure both real-time processing and detection accuracy within low-power embedded environments.

Among AI accelerators designed for embedded environments, the Hailo-8L supports computational processing in low-power settings. However, it is limited to 4D input tensors, making it difficult to directly apply 3D CNN architecture that requires 5D inputs. Consequently, the Hailo-8L has primarily been utilized for 2D image-based networks (2) and PointPillars based LiDAR detection models (3). In this study, we address these limitations and present the first on-chip implementation of 4D Radar-based 3D object detection model on the Hailo-8L AI accelerator. To this end, we employ the Radar Tensor Network with Height (RTNH) model (4) based on a 3D CNN architecture and convert the input tensor from a 5D structure to a 4D structure during the compilation process, thereby overcoming the input dimensionality constraint of the Hailo platform while maintaining the original model structure to enable computational processing. The main contributions of this paper are summarized as follows:

- To the best of our knowledge, this work presents a pioneering implementation of a 4D Radar-based 3D object detection model on the Hailo-8L AI accelerator.
- Proposal of an input tensor transformation method during the Hailo compilation process, enabling the deployment of 3D CNN architecture in embedded environments.
- Achievement of an $AP_{3D}$ of 46.47% and an $AP_{BEV}$ of 52.75%, maintaining comparable detection accuracy to GPU-based implementations while achieving an inference speed of 13.76 Hz.

## METHODOLOGY

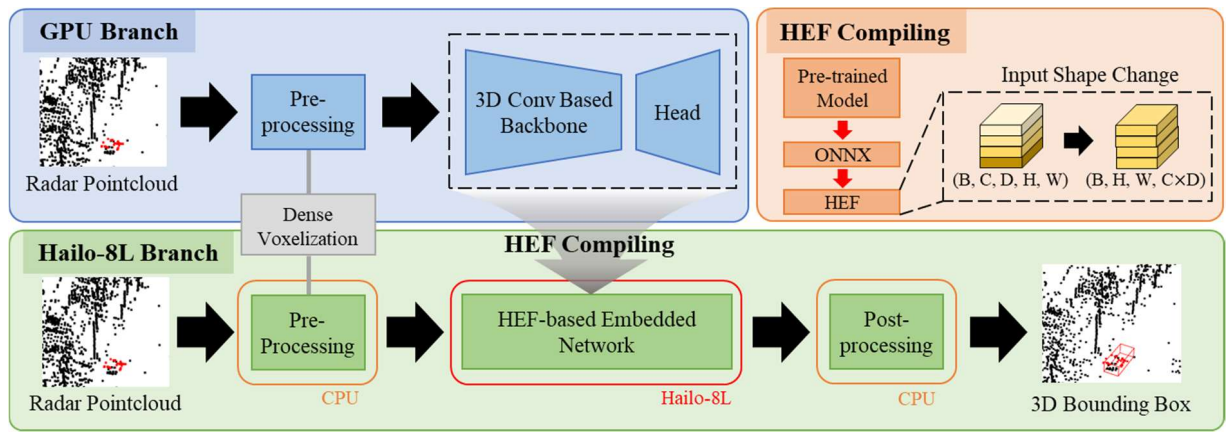

**Fig. 1. On-chip pipeline for 4D Radar-based 3D object detection on Hailo-8L.** The process consists of four main stages: HEF compiling, CPU-based pre-processing, on-chip inference on the Hailo-8L, and CPU-based post-processing. (B: batch size; C: channel; H: height; W: width; D: depth)

### HEF COMPILING

To integrate the RTNH model on the Hailo-8L, dense voxelization was applied, and the pre-trained model was converted into the ONNX format before being compiled into a Hailo Executable Format (HEF) file. During this process, since the Hailo-8L only supports 4D input tensors, the original 5D input structure (B, C, H, W, D) was reshaped into (B, H, W, C×D).

## PRE-PROCESSING ON HAILO-8L BRANCH

In the Hailo-8L branch, the same pre-processing procedures as those used in the GPU-based pipeline were applied. Point cloud voxelization was performed on the CPU, and the resulting voxel tensors were reshaped into the (B, H, W, C×D) format to meet the input tensor requirements for HEF compilation before being passed to the Hailo-8L accelerator.

## HEF-BASED EMBEDDED NETWORK

The compiled HEF file was deployed to the Hailo-8L accelerator using the VStream library provided by Hailo. The pre-processed input tensors were directly fed into the deployed model, preserving the 3D spatial structure during inference.

## POST-PROCESSING ON HAILO-8L BRANCH

Post-processing was performed on the CPU after inference on the Hailo-8L accelerator. The anchor generation method from the original RTNH model was used in the post-processing stage. A two-stage Non-Maximum Suppression (NMS) method was applied to improve computational efficiency and maintain detection accuracy: first, Axis-Aligned Bounding Box NMS was performed to filter redundant candidates, and then rotated BEV NMS based on polygonal IoU was applied to the remaining boxes.

# EXPERIMENTS

4D Radar point cloud dataset, self-collected using two radar sensors, was utilized for this study. The experiments were performed on Hailo-8L and NVIDIA GeForce RTX 3070.

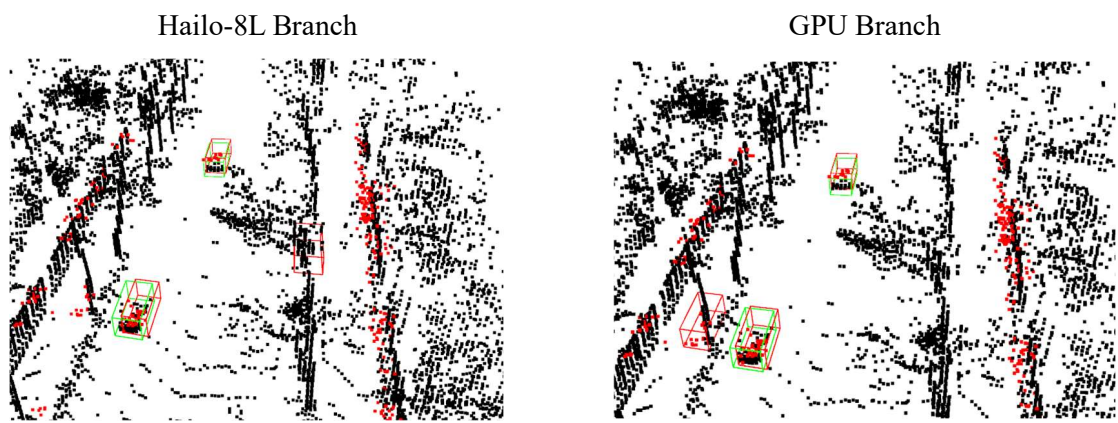

**Fig. 2. 3D object detection results: Hailo-8L output (left) and GPU output (right). Black points denote LiDAR point clouds, red points denote radar point clouds. Predicted boxes are shown in red, and ground truth boxes in green.**

| Hardware | $AP_{3D}$ [%] | $AP_{BEV}$ [%] | FPS[Hz] |
|---|---|---|---|
| Hailo-8L | 46.47 | **52.75 (+2.44)** | **13.76 (+0.85)** |
| GPU | **49.65 (+3.18)** | 50.31 | 12.91 |

**Table. 1. Quantitative Performance comparison between Hailo-8L and GPU**

Fig. 2 and Table 1 present the qualitative and quantitative comparison results, respectively. As shown in Fig. 2, the Hailo-8L achieves object detection performance comparable to that of the GPU. According to Table 1, the Hailo-8L shows a performance gap of 3.18% in 3D detection accuracy compared to the GPU, while achieving slightly higher performance in BEV detection. In addition, the Hailo-8L attains a faster inference speed than the GPU, maintaining both detection accuracy and inference speed in a low-power embedded environment.

|  | Pre-processing [CPU] | Network [Hailo-8L] | Post-processing [CPU] |
|---|---|---|---|
| Inference time | 0.017s | 0.028s | 0.022s |

**Table. 2. Inference time for Pre-processing, Hailo-8L based network, and Post-processing.**

According to Table 2, the Hailo-8L pipeline consists of pre-processing on the CPU (0.017s), inference on the Hailo-8L (0.028s), and post-processing on the CPU (0.022s). The total processing time per frame is approximately 0.067 seconds, corresponding to a theoretical 15 FPS, but the actual FPS drops to 13.76 due to CPU - Hailo-8L bottlenecks.

## ACKNOWLEDGEMENT
This work was supported by the National Research Foundation of Korea (NRF) grant funded by the Korea government (MSIT) (No. 2021R1A2C3008370). 4D Radar sensors used in this research were provided by Bitsensing Inc.